\documentclass[conference]{IEEEtran}
\IEEEoverridecommandlockouts
\usepackage{cite}
\usepackage{amsmath,amssymb,amsfonts}
\usepackage{algorithmic}
\usepackage{graphicx}
\usepackage{subcaption} 
\usepackage{textcomp}
\usepackage{xcolor}
\usepackage{bm}
\usepackage{booktabs}
\usepackage{hyperref}
\usepackage[acronym,toc,nopostdot]{glossaries}
\usepackage{placeins}
\usepackage{multirow}
\usepackage{makecell}
\makeglossaries
\newacronym{radarmot}{RadarMOT}{Radar-Informed 3D Multi-Object Tracking}
\newacronym{3dmot}{3D MOT}{3D multi-object tracking}
\newacronym{amota}{AMOTA}{Average Multi-Object Tracking Accuracy}
\newacronym{ids}{IDS}{Identity Switches}
\newacronym{truckscenes}{TruckScenes}{MAN-TruckScenes dataset}
\newacronym{kf}{KF}{Kalman filter}
\newacronym{bev}{BEV}{bird’s-eye view}
\newacronym{tbd}{TBD}{Tracking-by-Detection}
\newacronym{tba}{TBA}{Tracking-by-Attention}
\newacronym{iou}{IoU}{Intersection over Union}
\newacronym{nms}{NMS}{non-maximum suppression}

\def\BibTeX{{\rm B\kern-.05em{\sc i\kern-.025em b}\kern-.08em
    T\kern-.1667em\lower.7ex\hbox{E}\kern-.125emX}}
\begin{document}

\title{Radar-Informed 3D Multi-Object Tracking under Adverse Conditions\\}

\author{%
\IEEEauthorblockN{Bingxue Xu\textsuperscript{1}}
\and
\IEEEauthorblockN{Emil Hedemalm}
\and
\IEEEauthorblockN{Ajinkya Khoche\textsuperscript{1}}
\and
\IEEEauthorblockN{Patric Jensfelt\textsuperscript{1}}
\thanks{\textsuperscript{1}KTH Royal Institute of Technology, Stockholm 10044, Sweden.
Corresponding author’s e-mail: bingxue@kth.se.}%
}




\maketitle

\begin{abstract}
The challenge of 3D multi-object tracking is achieving robustness in real-world applications, for example under adverse conditions and maintaining consistency as distance increases. 
To overcome these challenges, sensor fusion approaches that combine LiDAR, cameras, and radar have emerged.
However, existing multimodal methods usually treat radar as another learned feature inside the network. When the overall model degrades in difficult environments, the robustness advantages that radar could provide are also reduced.
In this paper we propose RadarMOT, a radar-informed 3D multi-object tracking framework that explicitly uses radar point clouds as additional observations to refine state estimation and recover objects missed by the detector at long ranges.
Evaluations on the MAN-TruckScenes dataset show that RadarMOT consistently improves the
Average Multi-Object Tracking Accuracy (AMOTA) 
by 12.7\% at long range and up to 10.3\% in adverse weather. The code will be available at \url{https://github.com/bingxue-xu/radarmot}
\end{abstract}


\section{Introduction}
Safe 3D perception requires a system to continuously understand \emph{what} is around it
and \emph{where it is going} before acting. This is the task of \gls{3dmot}: maintaining identities and motion trajectories of detected objects across time. Trajectory prediction and motion planning depend directly on the
quality of these estimates. \gls{3dmot} operates by detecting objects independently in each frame and then associating them across time to form trajectories.
This makes \gls{3dmot} performance fundamentally dependent on two things:
(i) \emph{how accurately motion is estimated} (state estimation accuracy),
and (ii) \emph{how reliably detections are matched with tracked objects} (data association robustness).

LiDAR and cameras are the dominant perception sensors and have driven rapid progress in 3D detection and \gls{3dmot}~\cite{yin_center-based_2021, jiang_far3d_2023, liu_petr_2022, wang_exploring_2023, wang_detr3d_2021}.
However, both struggle at long range and under adverse conditions. 
LiDAR point clouds become sparse at long range and degrade in fog, rain, and snow.
Camera-based detectors show larger 3D depth errors at longer ranges\cite{jiang_far3d_2023} and are sensitive to light\cite{huang_l4dr_2024}.
On the other hand, radar is a motion-rich but geometry-poor sensor. Radar directly measures objects’ radial velocity via the Doppler effect, providing instantaneous motion without frame-to-frame differencing. It also works reliably in poor weather, low light, and at long range~\cite{huang_l4dr_2024}, making it a good complement to LiDAR and cameras where they are weakest.

Recent radar fusion methods~\cite{nabatiCenterFusionCenterbasedRadar2021, wolters_unleashing_2024, leiHVDetFusionSimpleRobust2023, zhangMixedFusionEfficientMultimodal2025} tend toward data-driven, mostly early-fusion strategies that learn radar features within the detection network. This is mainly due to: 1) detection primarily determining overall tracking performance, and 2) radar point clouds are sparse and noisy. Multimodal fusion methods~\cite{baumann_cr3dt_2024, li_farfusion_2024, wolters_unleashing_2024, woltersSpaRCSparseRadarCamera2024} explicitly use radar-refined detection velocity in data association to improve tracking performance. 
Despite achieving strong results on tracking benchmarks, these feature-fusion approaches still depend primarily on geometric cues from LiDAR or semantic cues from images. When these two sensors struggle under adverse weather, the fused representations also become unreliable, and the models undermine radar’s inherent advantages.

To reduce the computation burden on the GPU, we revisit both the radar sensing mechanism and Kalman filter-based \gls{3dmot} frameworks~\cite{weng_ab3dmot_2020, wangMCTrackUnified3D2024, chiu_probabilistic_2020, pang_simpletrack_2021, zhang_easy-poly_2026}. We first incorporate radar point clouds as an additional measurement in the Kalman filter update, which consistently boosts \gls{amota} and lowers false positives, indicating a more accurate state estimation. Next, we address motion blur during data preprocessing by leveraging radial velocity. Based on the observation that the distance-based tracker CenterPoint~\cite{yin_center-based_2021} yields more true positives than MCTrack~\cite{wangMCTrackUnified3D2024}, we further extend data association into a two-stage scheme that utilizes bidirectional distances with radar association, thereby stabilizing the tracking system.

Our proposed method RadarMOT achieves strong and consistent improvements on the \gls{truckscenes}, demonstrating that leveraging radar physical measurements in the tracking stage can substantially increase robustness when LiDAR and cameras degrade. Overall, RadarMOT improves \gls{amota} by +6.7\% over the MCTrack baseline and reduces \gls{ids} by 30\%, with the largest gains appearing at long range and under adverse conditions.
The contributions of this paper are:
\begin{itemize}
\item We propose \textbf{RadarMOT}, a radar-informed \gls{3dmot} framework that integrates radar as an explicit observation for tracking without deep learning, demostrates robustness under adverse conditions and at long ranges.
\item We introduce a practical \textbf{motion compensation} pipeline for multi-sweep radar aggregation that accounts for both ego motion and target motion, compensates temporal offsets using radar Doppler measurements.
\item We formulate a \textbf{radar-informed Kalman filter update} that uses associated radar radial velocities to refine tracked objects’ velocity and stabilize trajectories.
\item We propose a \textbf{two-stage association} strategy with cross-check association followed by radar association to reduce mismatches and recover detector misses, particularly at long range.
\end{itemize}

\section{RELATED WORK}\label{sec:relatedwork}
\gls{3dmot} is dominated by the geometry-based \gls{tbd} paradigm and the learning-based \gls{tba} paradigm. Most radar-based multimodal fusion approaches fall within the \gls{tba} category.

\noindent\textbf{Tracking by Detection}\label{sec:tbd}
\gls{tbd} divides the tracking task into four independently modules: detection, motion model, data association, and lifecycle management~\cite{pang_simpletrack_2021}. AB3DMOT~\cite{weng_ab3dmot_2020} established a strong baseline by coupling a Kalman filter~\cite{ramachandra_kalman_2018} with \gls{iou}-based Hungarian matching~\cite{hungarian_algorithm}.
The following methods have progressively refined the association strategy, from generalized \gls{iou} to normalization~\cite{zhengDistanceIoULossFaster2020}, lifting it to 3D~\cite{pang_simpletrack_2021}, and combining distance with ~\gls{iou}~\cite{wangMCTrackUnified3D2024}.
Another representative approach, CenterPoint~\cite{yin_center-based_2021}, replaces ~\gls{iou} with the Euclidean distance combined with greedy matching, achieving strong performance on nuScenes~\cite{caesar_nuscenes_2020} and Waymo~\cite{sunScalabilityPerceptionAutonomous2020}. \cite{chiu_probabilistic_2020} replaces the Euclidean distance with the Mahalanobis distance.
Other methods~\cite{liPolyMOTPolyhedralFramework2023, liFastPolyFastPolyhedral2024, zhang_easy-poly_2026, liu_imm-mot_2025} enhanced the motion model within the Kalman filter. SimpleTrack~\cite{pang_simpletrack_2021} proposes a two-stage association.
MCTrack~\cite{wangMCTrackUnified3D2024}, consolidates these advances into a unified and general framework that refines the Kalman filter and incorporates both ~\gls{iou} and distance into the data association stage, achieving state-of-the-art performance across the KITTI~\cite{geiger_are_2012}, nuScenes~\cite{caesar_nuscenes_2020}, and Waymo~\cite{sunScalabilityPerceptionAutonomous2020} datasets. We therefore adopt MCTrack as our baseline.

\noindent\textbf{Tracking by Attention}\label{sec:tba}
Emerging \gls{tba} approaches replace explicit associations with learned queries that carry object identity across frames. 
Recent \gls{3dmot} methods~\cite{meinhardt_trackformer_2022,zhangMUTR3DMulticameraTracking2022,pang_standing_2023,fischerCC3DTPanoramic3D2022,baumann_cr3dt_2024,leeDINOMOT3DMultiObject2025,zhangMotionTrackEndtoEndTransformerbased2023} encode tracked objects as queries, which attend to features of the current frame to both update object states and detect new objects in a single step,
eliminating the need for hand-crafted cost metrics or lifecycle heuristics.
Early approaches~\cite{meinhardt_trackformer_2022,sun_transtrack_2021} introduced the joint detection-and-tracking framework by propagating object queries over time and preserving identities via attention. Following works~\cite{zeng_motr_2022,yu_motrv3_2023} extended this idea to longer temporal modelling through track-aware supervision and improved training schemes.
Subsequent works~\cite{zeng_motr_2022, zhang_motrv2_2023, yu_motrv3_2023} skip the intermediate output detection and introduce end-to-end tracking frameworks.
Despite these advances, a common weakness of \gls{tba} methods is tight coupling. If the detector or features degrade (bad weather, night, long range, sparse points), tracking can degrade too, and changing sensors usually means retraining.

\noindent\textbf{Radar-based Tracking}\label{sec:radartracking}
Radar has recently regained attention due to its robustness to adverse weather, particularly in the scope of camera-radar fusion~\cite{yao_exploring_2024,shi2025radar}.
A dominant direction incorporates radar at the detection stage before tracking~\cite{nabatiCenterFusionCenterbasedRadar2021,nabatiCFTrackCenterbasedRadar2021,leiHVDetFusionSimpleRobust2023,zhangMixedFusionEfficientMultimodal2025,wolters_unleashing_2024} and improves \gls{3dmot} performance by radar-enhanced detection.
References~\cite{baumann_cr3dt_2024, li_farfusion_2024, pan_ratrack_2024, li_farfusion_2024-1} train models using radar velocity as input and employ the predicted bounding box velocity in motion-aware association.
DopplerTrack~\cite{zeng_simpler_2025} revisits Doppler velocity for moving object tracking, highlighting the potential of radar for motion-centric tracking.
RaTrack~\cite{pan_ratrack_2024} presents a radar-only method for tracking moving objects that avoids relying on 3D bounding boxes and object classification by using a multi-task network.
Similarly, our approach performs motion segmentation, but we extract moving objects via geometric filtering, thus bypassing the need for deep learning.

\section{METHOD}
\begin{figure}[!htbp]
\centerline{\includegraphics[width=0.9\columnwidth]{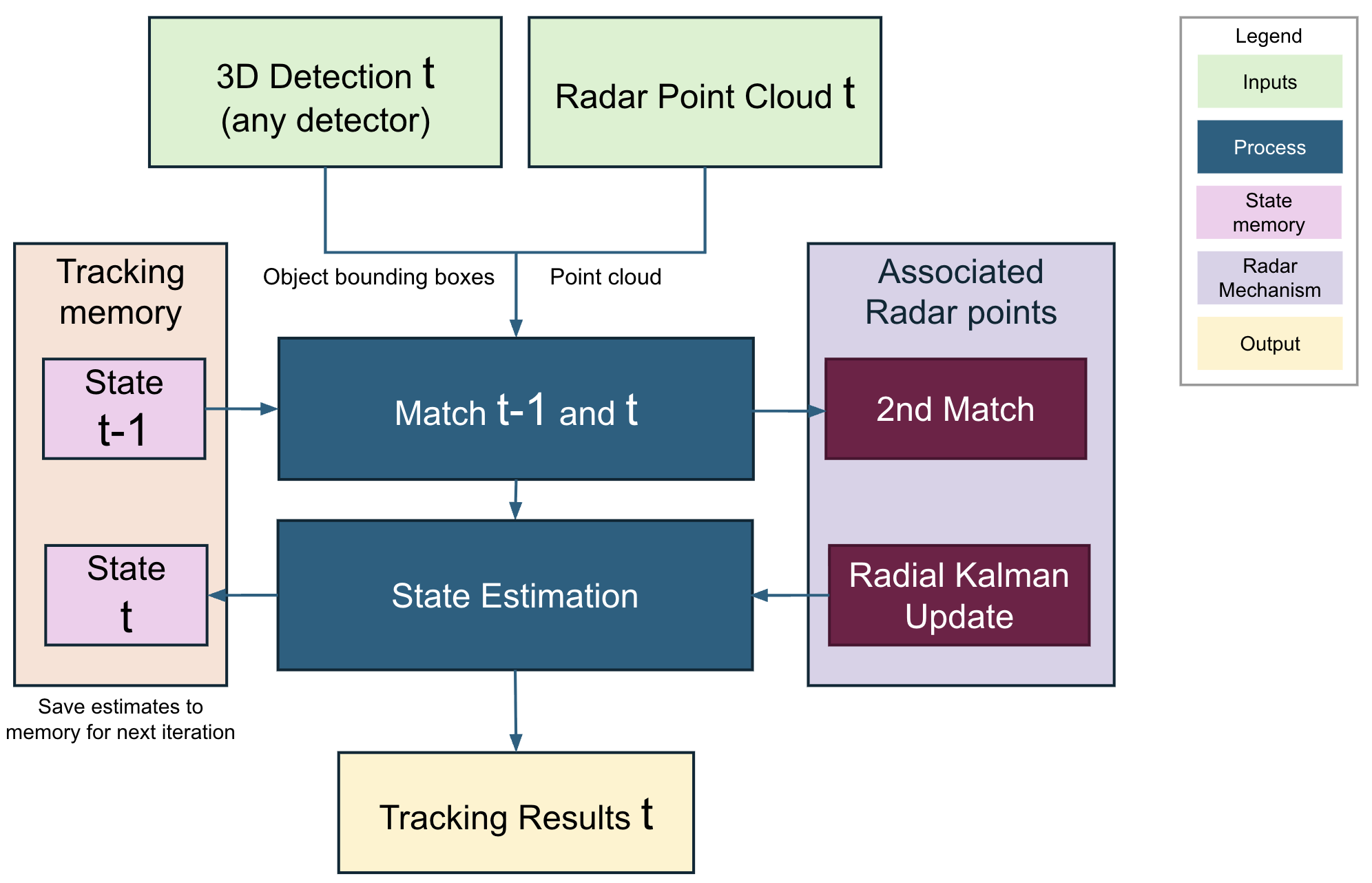}}
\caption{Overall framework of the proposed RadarMOT, built
upon MCTrack~\cite{wangMCTrackUnified3D2024} and incorporating radar. It first associates tracked objects from both detection and radar in two stages, and then refines the state estimate by a radial velocity-constrained Kalman filter.}
\label{fig:radarmotframework}
\end{figure}

We propose RadarMOT, as shown in Fig.~\ref{fig:radarmotframework}. It builds upon MCTrack \cite{wangMCTrackUnified3D2024} and incorporates radar data. We treat radar data as additional observations alongside 3D detection bounding boxes. The radial velocity is used to refine state estimation in the Kalman filter, and the radar points indicate object presence when objects are temporarily missed by the detector. This direct design increases the robustness of tracking when detection is weak and compatible with any detector. 

\subsection{Motion Compensation} \label{sec:compensation}
We observed the motion blur problem caused by the ego motion and target motion while aggregating multi-sweep radar data, and addressed this displacement using both ego motion and radar radial velocity.

\noindent\textbf{Ego motion compensation}
Following the ego-motion compensation~\cite{fentMANTruckScenesMultimodal2024}, we observe that static objects still have residual radial velocities when the vehicle rotate. Therefore we also account for the ego rotation. By doing so, we correctly remove both the translational and rotational ego-motion effects.
Let \(i\) denote the $i$th radar point, \(\bm{b}_i\) denotes the line-of-sight unit vector of radar point \(p_i\).
After ego motion compensation, the point velocity in the reference coordinate system is written as:
\begin{equation}
\bm{v}_{\mathrm{comp}}^{(i)} 
= \bm{v}_{\mathrm{rel}}^{(i)} + \bm{v}_{\mathrm{ego}} + \bm{\omega}_{\mathrm{ego}} \times \bm{r}^{(i)}
\label{eq:egocompensation}
\end{equation}
where $\bm{v}_{\mathrm{rel}}^{(i)}$ denotes the radar-reported relative velocity of point $i$, $\bm{v}_{\mathrm{ego}}$ is the ego linear velocity, $\bm{\omega}_{\mathrm{ego}}$ is the ego angular velocity, and $\bm{r}^{(i)}$ is the vector from the ego rotation center to the radar point. 

\noindent\textbf{Radar motion compensation} 
After ego-motion compensation, dynamic points can still appear displaced because radar points are acquired at the sweep timestamp rather than at the keyframe timestamp. Unlike LiDAR methods that infer object motion via scene flow \cite{zhang2025himo}, radar Doppler directly measures velocity, capturing non-ego motion without learning. To mitigate the temporal offset, each radar point is translated according to its estimated motion over the time interval between the sweep timestamp $t_s$ and the keyframe timestamp $t_k$. The resulting motion-compensated point position is given by:
\begin{equation}
\bm{p}_{\mathrm{comp}}^{(i)}
\approx
\bm{p}_{k}^{(i)}
+
\Delta r^{(i)} \bm{b}_i^{(i)}
\label{eq:radarcompensation}
\end{equation}
where $\bm{p}_{k}^{(i)}$ denotes the point position after transformation to the keyframe in the reference coordinate system, $\bm{b}_i^{(i)}$ is the radar line-of-sight unit vector, and $\Delta r^{(i)} = v_{\mathrm{radial}}^{(i)} (t_k - t_s)$ is the estimated radial displacement during the time offset. 

\begin{figure}[!htbp]
    \centerline{\includegraphics[width=\columnwidth]{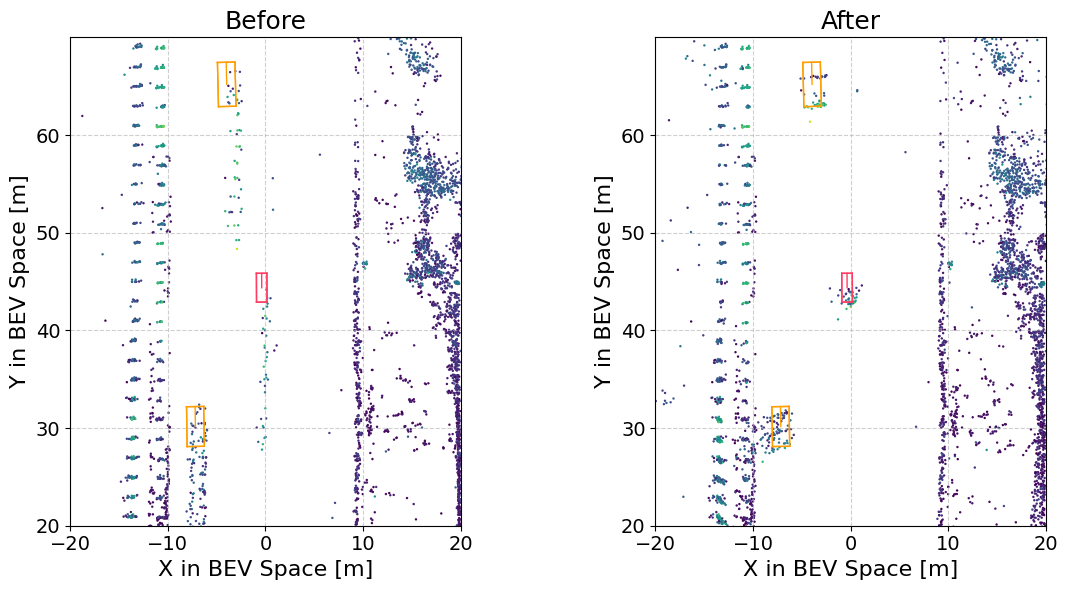}}
    \caption[Motion compensation effect]{Comparison of radar point clouds before and after motion compensation. Bounding boxes indicate annotated ground truth.}
    \label{fig:motioncompensation_zoomin}
\end{figure}

Fig.~\ref{fig:motioncompensation_zoomin} shows the effect of the proposed motion compensation in the \gls{bev}. Because radar measures only radial velocity, tangential velocity is unobservable, causing residual distortion, especially for objects crossing the radar’s line of sight. This effect is strongest near the ego vehicle, so a nearby region is discarded (15 m ego-centric radius in this implementation).

\subsection{Radar Informed Kalman filter}\label{sec:radarkf}
To update state estimation with radar information, we treat the associated radar radial velocity as an additional observation in a Kalman filter, thereby refining the tracked objects’ planar velocity and indirectly stabilizing their positions over time.

In short, for each tracked object we (i) associate radar points that fall inside an inflated track box and have consistent line-of-sight velocity with the box motion~\eqref{eq:radarassoc}, (ii) build an observation matrix \(\bm{H}_{r,k}\) whose rows project the track’s planar velocity onto each radar point’s line-of-sight direction ~\eqref{eq:bearing}\eqref{eq:Hrk}, and (iii) update the state with the innovation between measured and predicted radial velocities ~\eqref{eq:radar_kf_update}. This uses radar primarily to constrain velocity, which improves tracking robustness under occlusion and reduces velocity drift.

\noindent\textbf{Motion Model}  
We use a constant-velocity motion model, Let \(k\) denote the keyframe index. The state
\(\mathbf{x}_k = [x_k, y_k, v_{x,k}, v_{y,k}]^\top\)
denotes the object position and planar velocity in the global coordinate system. 
$F$ denotes the state transition matrix, \(\bm{Q}_k\) denotes the process noise. The predicted state is then given as:
\begin{equation}
\bm{x}_{k|k-1} = \bm{F}\bm{x}_{k-1|k-1} + \bm{w}_k 
\qquad
\bm{w}_k \sim \mathcal{N}(\bm{0}, \bm{Q}_k)\
\label{eq:motionmodel}
\end{equation}

\noindent\textbf{Radar Observation Model}
To handle noisy radar points, only radar points likely to be reflecting from the 
same object are used as observations. Let \(p_i\) denote the \(i\)-th radar point, and let \(\bm{p}^{(i)}_s \in \mathbb{R}^3\) denote its position, with measured radial velocity \(v^{(i)}_{\mathrm{radar}} 
\in \mathbb{R}\), and let \(v^{(i)}_{\mathrm{box}} \in \mathbb{R}\) 
denote the bounding box velocity projected onto the line-of-sight of 
point \(i\). The associated radar point set is defined by two filters:
\begin{equation}
\mathcal{R}_{\mathrm{assoc}} = \bigl\{\, p_i \in \mathcal{R}_k
\;\big|\;
\bm{p}^{(i)}_g \in \alpha\mathcal{B}^t_k
\;\wedge\;
|v^{(i)}_{\mathrm{radar}} - v^{(i)}_{\mathrm{box}}| \leq \delta_v
\,\bigr\}
\label{eq:radarassoc}
\end{equation}
where 
\begin{itemize}
    \item $\alpha\mathcal{B}_k$ restricts radar points to the track bounding box in XY plane, $\alpha$ denotes the inflation rate.
    \item $\delta_v$ removes outliers based on the radial velocity difference between each radar point $v^{(i)}_{\mathrm{radar}}$ and the bounding box velocity $v^{(i)}_{\mathrm{box}}$ projected onto the radar point’s line-of-sight direction.
\end{itemize}

To construct the radar observation matrix \(\bm{H}_{r,k}\), where $r$ means observation from radar at the \(k\)-th frame, the line-of-sight 
unit vector \(\bm{b}_i\) of each associated radar point \(i\) is defined in the radar sensor 
coordinate system \(S_i\) as:
\begin{equation}
    \bm{b}_i = \frac{\bm{p}_s^{(i)}}{\left\|\bm{p}_s^{(i)}\right\|}
    \in \mathbb{R}^3
    \label{eq:bearing}
\end{equation}
The observation function \(h_i\) for point \(i\) projects the track velocity 
onto this line-of-sight direction, defined as:
\begin{equation}
    h_i(\bm{x}_k) = \bm{h}_i^\top \bm{x}_k, \qquad 
    \bm{h}_i^\top = 
    \begin{bmatrix} 0 & 0 & \bm{b}_i^\top\,{}^{S_i}\bm{R}_{\mathrm{global},xy} \end{bmatrix}
    \in \mathbb{R}^{1 \times 4}
    \label{eq:observationfunction}
\end{equation}
where \({}^{S_i}\bm{R}_{\mathrm{global},xy} \in 
\mathbb{R}^{2 \times 3}\) extracts the planar rows of the 
global-to-radar rotation, and the position columns of 
\(\bm{h}_i^\top\) are zero since only velocity enters the projection.
Stacking \(M\) such rows gives the joint observation matrix:
\begin{equation}
    \bm{H}_{r,k} = 
    \begin{bmatrix} \bm{h}_1^\top \\ \vdots \\ \bm{h}_M^\top 
    \end{bmatrix} \in \mathbb{R}^{M \times 4}
    \label{eq:Hrk}
\end{equation}

Let \(\bm{z}^r_k = [z_1, z_2, \ldots, z_M]^\top \in \mathbb{R}^M\) 
denote the vector of radial velocities measured by the \(M\) 
associated radar points. Following the standard linear observation 
model~\cite{thrun2005probabilistic}, the joint radar observation model is as:
\begin{equation}
    \bm{z}^r_k = \bm{H}_{r,k}\bm{x}_k + \bm{n}_k, \qquad 
    \bm{n}_k \sim \mathcal{N}(\bm{0}, \bm{R}_{r,k})
    \label{eq:radar_multi_obs}
\end{equation}
where \(\bm{x}_k\) is the true object state, and \(\bm{R}_{r,k} \in 
\mathbb{R}^{M \times M}\) denotes the measurement noise covariance.

\begin{figure}[!htbp]
\centerline{\includegraphics[width=0.75\columnwidth]{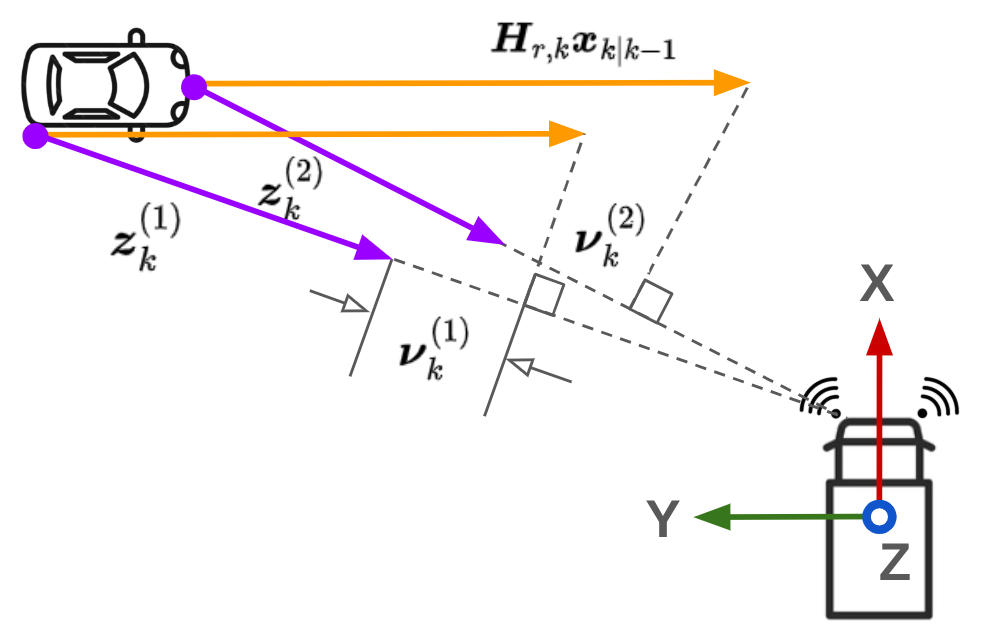}}
\caption{Radar-informed Kalman update. Purple points and arrows represent radar points and motion-compensated velocity. Orange arrows represent the predicted velocity coming from the bounding box.}
\label{fig:radarkfupdate}
\end{figure}

\noindent\textbf{Radar-Informed Kalman Filter Update} 
Given the radar observation, the tracking state is then updated using a Kalman filter, as illustrated in Fig.~\ref{fig:radarkfupdate}, the radar-informed state is updated by:
\begin{equation}
\begin{aligned}
\bm{\nu}_k &= \bm{z}^r_k - \bm{H}_{r,k}\bm{x}_{k|k-1} \\
\bm{x}_{k|k} &= \bm{x}_{k|k-1} + \bm{K}_k\bm{\nu}_k
\label{eq:radar_kf_update}    
\end{aligned}
\end{equation}
where
\begin{itemize}
    \item $\bm{\nu}_k$ denote the innovation derived from the difference between the observed radial velocity and the motion model predicted radial velocity.
    \item $\bm{z}^r_k$ denotes the observed radial velocity.
    \item $\bm{H}_{r,k}\bm{x}_{k|k-1}$ denotes the track state estimate predicted radial velocity.
    \item $\bm{x}_{k|k-1}$ and $\bm{x}_{k|k}$ denote the predicted and updated track states.
    \item $\bm{K}_k$ is the Kalman gain.
\end{itemize}
\subsection{Two-Stage Association}
To reduce mismatches, which are the primary cause of identity switches in tracked objects, we propose a two-stage association that first performs a bidirectional cross-check association and then applies a radar association.

\noindent\textbf{Cross-Check Association}
Instead of associating at a single timestamp (e.g., either comparing the forward prediction with the detection at time $t$, or comparing the track with the backward predicted detection at time $t-1$), we use both forward and backward predictions for mutual cross-checking and combine a speed similarity to further minimize the influence of object orientation. The cost function is defined as:
\begin{equation}
cost =
\frac{\alpha \|\bm{p}_t^{trk}-\bm{p}_t^{det}\|}{v_t^{trk}\Delta{t}}
+
\frac{\beta\|\bm{p}_{t-1}^{trk}-\bm{p}_{t-1}^{det}\|}{v_t^{det}\Delta{t}}
+
\frac{\delta|v^{trk} - v^{det}|}{\max(v^{trk},v^{det})}
\label{eq:crosscheckcost}
\end{equation}
where
\begin{itemize}
    \item $\bm{p}_t^{trk}$ and $\bm{p}_t^{det}$ are the track and detection centers at time $t$.
    \item $\bm{p}_{t-1}^{trk}$ and $\bm{p}_{t-1}^{det}$ are the track center and backward-predicted detection center at time $t-1$.
    \item $v^{det}$ and $v^{trk}$ are the detection and track speeds, reducing the influence of orientation error.
    \item $\alpha$, $\beta$, and $\delta$ are weighting factors.
\end{itemize}

\noindent\textbf{Radar Association}
To recover the objects missed by the detector, unmatched tracked objects are associated with radar measurements in a second stage.

We first remove the points that are close to detections to avoid duplicate associations and remove stationary clutter, then select candidate radar points for unmatched tracked objects by position and velocity similarity.
\begin{equation}
\mathcal{R}^{t}_{\mathrm{assoc}} = \bigl\{\, p_i \in \mathcal{R}_k
\;\big|\;
p_i \in \beta\hat{\mathcal{B}}^t_k
\;\wedge\;
|v^{(i)}_{\mathrm{radar}} - v^{(i)}_{\mathrm{pred}}| \leq \delta_v
\,\bigr\}
\label{eq:radarmatching}
\end{equation}
where $\beta\hat{\mathcal{B}}^t_k$ restricts radar points to the track \emph{predicted} bounding box, $\beta$ denotes the inflation rate.
$\delta_v$ removes outliers based on the radial velocity difference between each radar point $v^{(i)}_{\mathrm{radar}}$ and the predicted track velocity $v^{(i)}_{\mathrm{pred}}$ projected onto the radar point's line-of-sight direction.

If more than $N_{\min}$ points are collected, the track is considered radar-seen for the current frame, and the radar-informed Kalman filter update~\eqref{eq:radar_kf_update} is applied to refine the state estimation.

\begin{table*}[!t]
    \caption[Performance comparison on MAN-TruckScenes validation set]{Performance comparison on MAN-TruckScenes validation set with the same detector. (Bic., Motor, Ped., Tra., Tru.) denote (Bicycle, Motorcycle, Pedestrian, Trailer, Truck). Bold indicates best performance per column.}
    \label{tab:mainresults}
    \renewcommand{\arraystretch}{1.1}
    \centering
    \resizebox{\textwidth}{!}{%
    \begin{tabular}{llcccccccccccc}
    \toprule
    \multirow[c]{2}{*}{\textbf{Method}} &
    \multirow[c]{2}{*}{\textbf{Detector}} &
    \multicolumn{8}{c}{\textbf{AMOTA\%\,$\uparrow$}} &
    \multirow[c]{2}{*}{\textbf{TP\,$\uparrow$}} &
    \multirow[c]{2}{*}{\textbf{FP\,$\downarrow$}} &
    \multirow[c]{2}{*}{\textbf{FN\,$\downarrow$}} &
    \multirow[c]{2}{*}{\textbf{IDS\,$\downarrow$}} \\
    \cmidrule(lr){3-10}
    & &
    \textbf{Overall} &
    \textbf{Bic.} &
    \textbf{Bus} &
    \textbf{Car} &
    \textbf{Motor} &
    \textbf{Ped.} &
    \textbf{Tra.} &
    \textbf{Tru.} &
    & & & \\
    \midrule
    CenterPoint~\cite{yin_center-based_2021}
      & CenterPoint~\cite{yin_center-based_2021}
      & 22.8 & 9.3 & 0.0 & 51.1 & 23.5 & 37.7 & 0.9 & 37.2
      & 37149 & \textbf{8467} & 26912 & 7278 \\
    MCTrack~\cite{wangMCTrackUnified3D2024}
      & CenterPoint~\cite{yin_center-based_2021}
      & 26.6 & 11.9 & 0.0 & 40.0 & \textbf{59.4} & 36.4 & 0.7 & 37.6
      & 33636 & 13026 & 32378 & 5325 \\
    RadarMOT (ours)
      & CenterPoint~\cite{yin_center-based_2021}
      & \textbf{33.3} & \textbf{17.0} & 0.0 & \textbf{53.6} & 56.2 & \textbf{46.5} & \textbf{13.0} & \textbf{46.8}
      & \textbf{42717} & 12257 & \textbf{24906} & \textbf{3716} \\
    \bottomrule
    \end{tabular}%
    }
\end{table*}

\section{EXPERIMENTS SETUP} \label{sec:experiments}
\noindent\textbf{Datasets} \label{sec:dataset}
Experiments are conducted on the largest public radar dataset \gls{truckscenes}~\cite{fentMANTruckScenesMultimodal2024} for autonomous driving, which includes 747 scenes of 20 seconds each, with keyframes at 2 Hz. The sensor suite includes 4 cameras, 6 LiDAR units, and 6 4D radar sensors with full 360-degree coverage. It covers diverse weather conditions (rain, fog, snow), lighting situations (darkness, glare, twilight), and driving areas (highway, urban, rural).
This dataset also reflects the dynamic, long-range perception and occlusion challenges: 40\% of annotated objects move faster than 20\,m/s,  50\% of all annotations are beyond 75 m, and 39\% of the objects have a visibility of less than 41\%~\cite{fentMANTruckScenesMultimodal2024}. Thus, evaluation is conducted using this dataset to investigate robustness.

\noindent\textbf{Evaluation Metrics}\label{sec:evaluationmetrics}
We follow the official nuScenes tracking challenge~\cite{nuscenes_tracking_metrics} and report \gls{amota}. 
For all metrics, true positives are obtained by applying a 2\,m threshold to the 2D center distance on the ground plane. 
\gls{amota} and MOTAR are defined as:
\begin{equation}
\mathrm{AMOTA} = \frac{1}{n - 1} \sum_{r \in \left\{ \frac{1}{n-1}, \frac{2}{n-1}, \dots, 1 \right\}} \mathrm{MOTAR}
\end{equation}
\begin{equation}
\mathrm{MOTAR} = \max \left( 0,\; 1 - \frac{\mathrm{IDS}_r + \mathrm{FP}_r + \mathrm{FN}_r - (1 - r)\cdot P}{r \cdot P} \right)
\end{equation}
where $n$ denotes the number of recall sampling points, and $r$ is the recall threshold. $P$ refers to the number of ground-truth positives for the current class. ($\mathrm{IDS}$,$\mathrm{FP}$,$\mathrm{FN}$) are the number of (identity switches, false positives, false negatives).

As this is the first \gls{3dmot} work on \gls{truckscenes}, the classes follow nuScenes tracking challenges\cite{nuscenes_tracking_metrics} and ranges are adopted from \gls{truckscenes} detection benchmark~\cite{fentMANTruckScenesMultimodal2024}, i.e., bigger vehicles (car, truck, bus, trailer) are evaluated up to 150 m and smaller objects (pedestrian, motorcycle, bicycle) up to 75 m. 

\noindent\textbf{Detection Input}\label{sec:detectionbaseline}
We use the same 3D detection results obtained by CenterPoint~\cite{yin_center-based_2021} on the \gls{truckscenes} validation dataset. The detection model is reproduced following the training recipe in TruckScenes~\cite{fentMANTruckScenesMultimodal2024}. It's worth mentioning that CenterPoint~\cite{yin_center-based_2021} achieves 58\% mean Average Precision (mAP) on nuScenes but only 33\% mAP on \gls{truckscenes} validation set, indicating that the environmental conditions in \gls{truckscenes} are more challenging for 3D perception. 

\noindent\textbf{Baseline}\label{sec:baselines}
We use MCTrack~\cite{wangMCTrackUnified3D2024} as baseline, as described in Section~\ref{sec:tbd}. We report the MCTrack tracking results on the \gls{truckscenes} validation set in the second row of Table~\ref{tab:mainresults}. Additionally, we report the CenterPoint tracker in the first row of Table~\ref{tab:mainresults}.

\section{Results} \label{sec:mainresults}
We report our results on the \gls{truckscenes} validation set in Table~\ref{tab:mainresults}.  
Our proposed method RadarMOT achieves 33.3\% in terms of overall \gls{amota}, an absolute improvement of 6.7\% over the MCTrack baseline, and a relative 30\% reduction in \gls{ids}.
We found that the distance-based association method is preferable in highly dynamic scenarios. 
We observed that CenterPoint achieves more true positives and fewer false negatives than MCTrack, but at the cost of a substantially increased number of \gls{ids}. Thanks to the bidirectional distance association and radar-refined state estimation, RadarMOT combines the strengths of both methods without introducing additional false positives.
To the best of our knowledge, this is the first \gls{3dmot} investigation on the \gls{truckscenes}, offering a conservative baseline for the community.

\begin{figure}[!htbp]
\centerline{\includegraphics[width=\columnwidth]{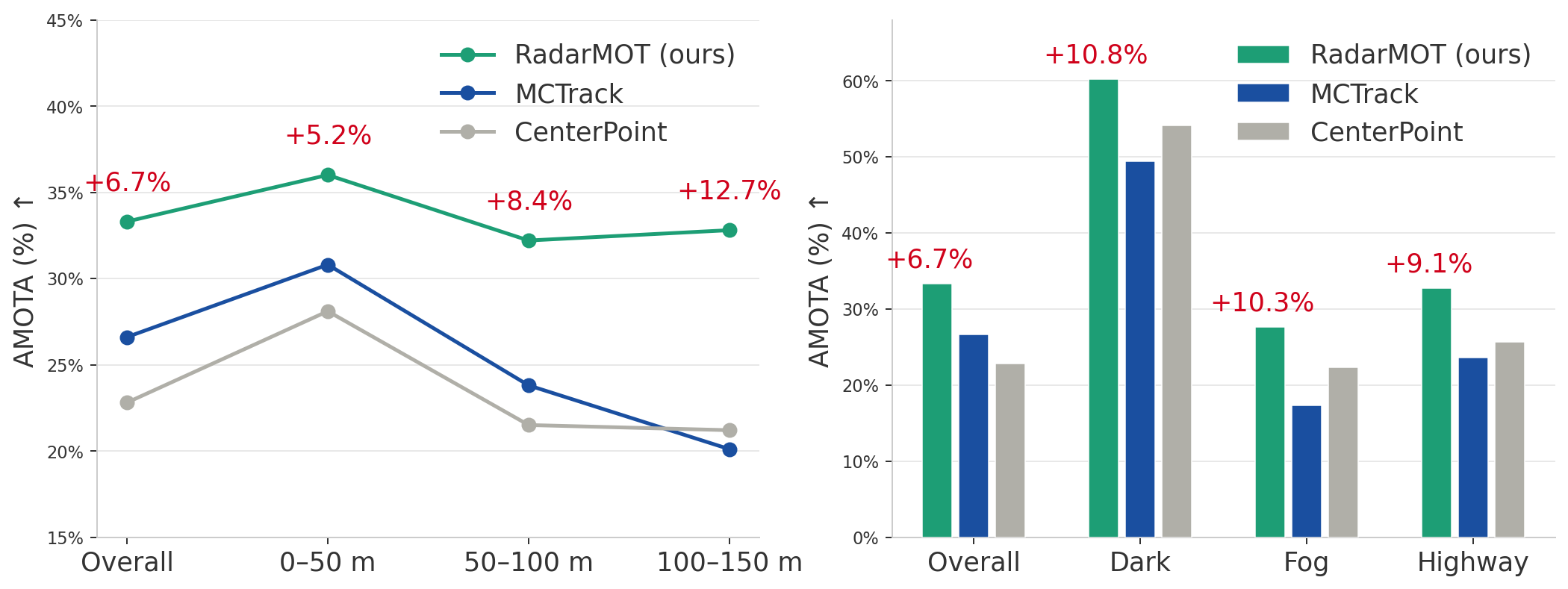}}
\caption{Comparison across range and adverse conditions on \gls{truckscenes} validation set.}
\label{fig:range_analysis}
\end{figure}

\noindent\textbf{Long Range} \label{sec:analysis_range}
To reveal how RadarMOT behaves as range increases, we evaluate the performance across three range bins up to 150\,m, as shown in Table~\ref{tab:rangeresult} and Fig.~\ref{fig:range_analysis}. 
The growing margin reflects radar’s increasing contribution as LiDAR point density drops with range. Our method achieves an absolute +12.7\% \gls{amota} improvement at the third bin (100–150 m) compared to baseline MCTrack.
\begin{table}[!htbp]
    \caption[Performance comparison across ranges]{Comparison across ranges on \gls{truckscenes} validation set. All methods use the same CenterPoint~\cite{yin_center-based_2021} detector. Bold indicates best value per column.}
    \label{tab:rangeresult}
    \renewcommand{\arraystretch}{1.1}
    \centering
    \resizebox{\columnwidth}{!}{%
    \begin{tabular}{lcccc}
    \toprule
    \multirow{2}{*}{\textbf{Method}} &
    \textbf{Overall} &
    \textbf{I: 0--50\,m} &
    \textbf{II: 50--100\,m} &
    \textbf{III: 100--150\,m} \\
    \cmidrule(lr){2-2}\cmidrule(lr){3-3}\cmidrule(lr){4-4}\cmidrule(lr){5-5}
    & \textbf{AMOTA\% / IDS}
      & \textbf{AMOTA\% / IDS}
      & \textbf{AMOTA\% / IDS}
      & \textbf{AMOTA\% / IDS} \\
    \midrule
    CenterPoint~\cite{yin_center-based_2021}
      &  22.8 / 7278 & 28.1 / 2766 & 21.5 / 3107 & 21.2 / 1518 \\
    MCTrack~\cite{wangMCTrackUnified3D2024}
      & 26.6 / 5325 & 30.8 / 1804 & 23.8 / 1704 & 20.1 / \textbf{836} \\
    RadarMOT (ours)
      & \textbf{33.3} / \textbf{3716} & \textbf{36.0} / \textbf{1321} & \textbf{32.2} / \textbf{1451} & \textbf{32.8} / 895 \\
    \bottomrule
    \end{tabular}
    }
\end{table}

\noindent\textbf{Adverse Conditions} \label{sec:analysis_conditions}
We also evaluate the performance under adverse conditions in different weather, lighting, and driving areas, see Table~\ref{tab:conditionresult} and Fig.~\ref{fig:range_analysis}. RadarMOT achieves consistent absolute AMOTA improvements over the baseline MCTrack: +10.3\% in fog, +10.8\% at night, and +9.1\% on highways. 
Contrary to our initial hypothesis, RadarMOT performs worse than the baseline in snow. 

\begin{table}[!htbp]
    \caption[Performance comparison in adverse conditions]{Comparison under adverse weather, lighting conditions and driving areas on \gls{truckscenes}
    validation set.
    Bold indicates the best value per row per metric.}
    \label{tab:conditionresult}
    \renewcommand{\arraystretch}{1.1}
    \centering
    \resizebox{\columnwidth}{!}{%
    \begin{tabular}{ll
        r@{\,/\,}r
        r@{\,/\,}r
        r@{\,/\,}r
    }
    \toprule
    \multirow{2}{*}{\textbf{Group}} &
    \multirow{2}{*}{\textbf{Condition}} &
    \multicolumn{2}{c}{\textbf{CenterPoint~\cite{yin_center-based_2021}}} &
    \multicolumn{2}{c}{\textbf{MCTrack~\cite{wangMCTrackUnified3D2024}}} &
    \multicolumn{2}{c}{\textbf{RadarMOT (ours)}} \\
    \cmidrule(lr){3-4}\cmidrule(lr){5-6}\cmidrule(lr){7-8}
    & &
    \textbf{AMOTA\%} & \textbf{IDS} &
    \textbf{AMOTA\%} & \textbf{IDS} &
    \textbf{AMOTA\%} & \textbf{IDS} \\
    \midrule
    \multirow{5}{*}{Weather}
      & Clear        & 36.0 & 3040 & 41.4 & 2454 & \textbf{46.0} & \textbf{1552} \\
      & Overcast     & 24.2 & 2753 & 29.0 & 1775 & \textbf{33.6} & \textbf{1307} \\
      & Rain         & 20.9 & 1458 & 18.3 &  839 & \textbf{26.1} & \textbf{771} \\
      & \textbf{Fog} & 22.3 &  173 & 17.3 &  109 & \textbf{27.6} & \textbf{76} \\
      & Snow         & 16.2 &   38 & \textbf{20.0} & 34 & 17.8 & \textbf{25} \\
    \midrule
    \multirow{4}{*}{Lighting}
      & \textbf{Dark} & 54.1 & 1388 & 49.4 & 674  & \textbf{60.2} & 728 \\
      & Illuminated   & 22.7 & 4627 & 33.0 & 2952 & \textbf{38.8} & \textbf{2317} \\
      & Twilight      & 25.8 & 1095 & 26.8 &  906 & \textbf{31.0} & \textbf{508} \\
      & Glare         & 16.3 &  494 & 13.2 &  361 & \textbf{16.8} & \textbf{219} \\
    \midrule
    \multirow{5}{*}{Area}
      & \textbf{Highway} & 25.7 & 5833 & 23.6 & 3975 & \textbf{32.7} & \textbf{2702} \\
      & City             & 21.5 &  887 & 24.4 &  687 & \textbf{27.4} & \textbf{577} \\
      & Rural            & 16.4 &  463 & 19.1 &  377 & \textbf{23.3} & \textbf{303} \\
      & Residential      & 25.3 &   53 & \textbf{29.6} & 83 & 28.6 & \textbf{54} \\
      & Terminal         & 51.9 &   33 & 56.5 & \textbf{9} & \textbf{56.6} & 13 \\
    \bottomrule
    \end{tabular}
    }
\end{table}

\begin{figure*}[!t]
    \centering
    \begin{subfigure}[b]{0.48\textwidth}
        \centering
        \includegraphics[width=\textwidth]{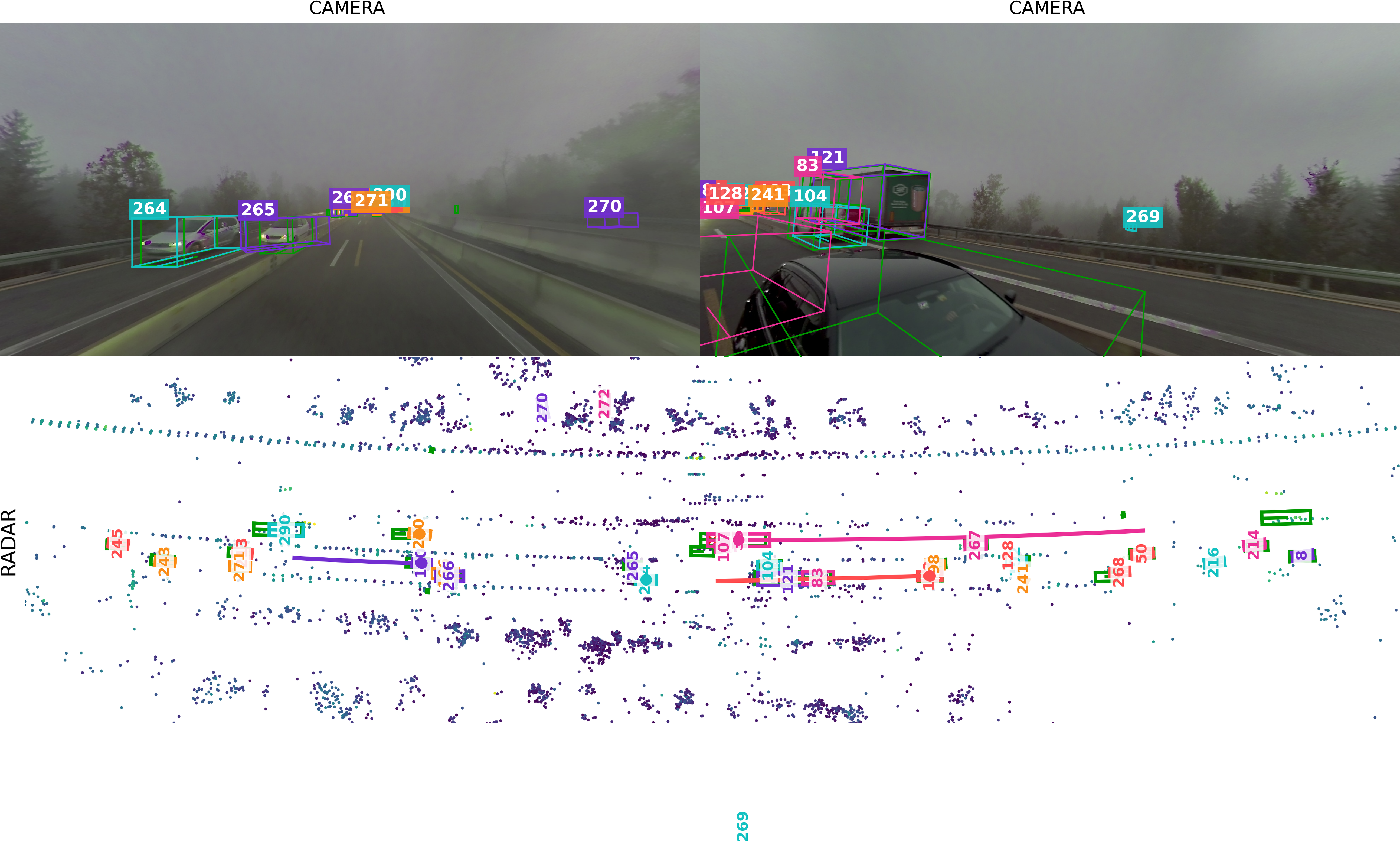}
        \caption[MCTrack (fog)]{MCTrack}
    \end{subfigure}
    \hfill
    \begin{subfigure}[b]{0.48\textwidth}
        \centering
        \includegraphics[width=\textwidth]{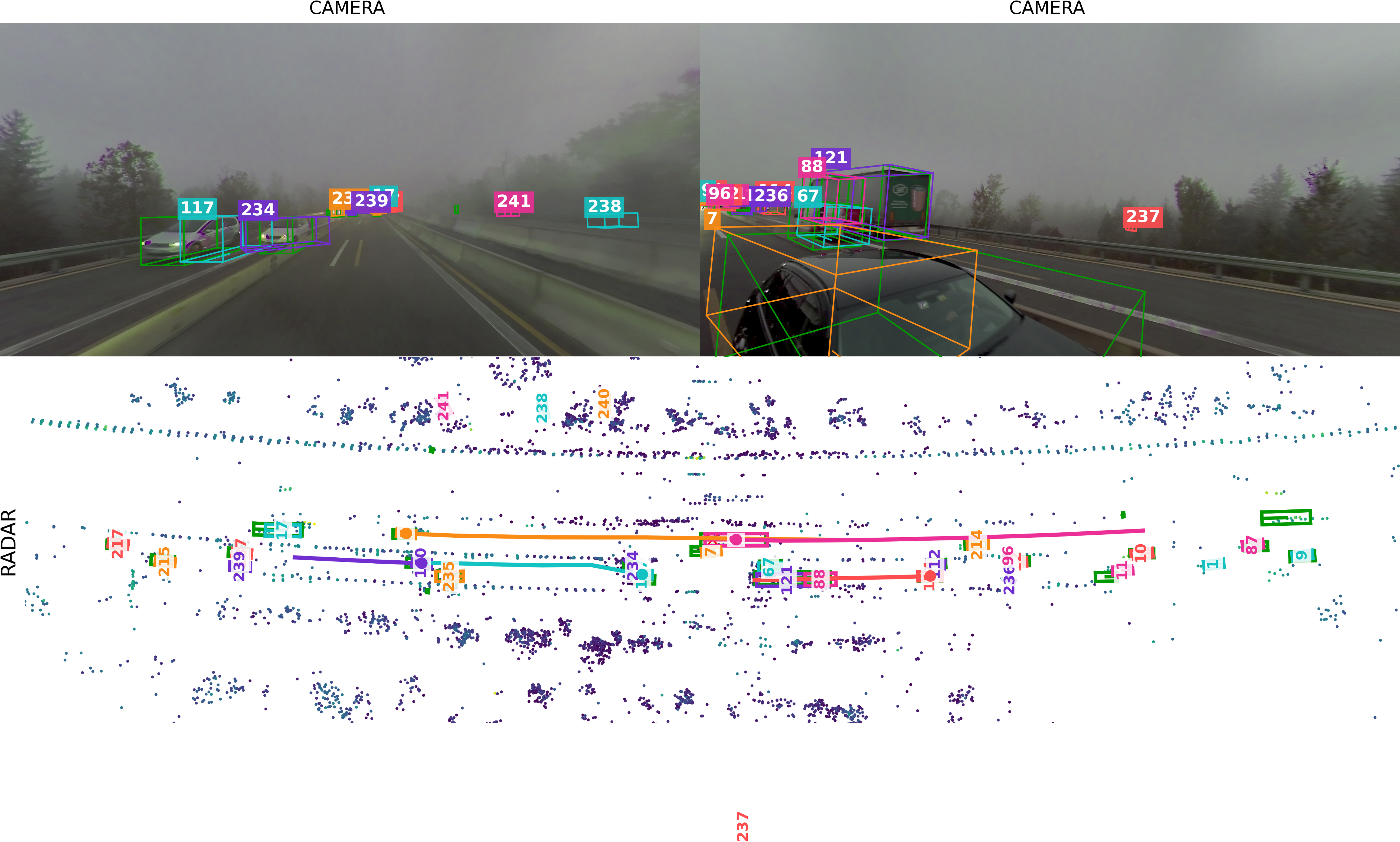}
        \caption[RadarMOT (fog)]{RadarMOT (ours)}
    \end{subfigure}
    \caption[Qualitative comparison]{Qualitative comparison on \gls{truckscenes} in fog and at long distances. Green bounding boxes indicate the ground truth, and same‑colored, numbered bounding boxes and trajectories represent individual tracking results. Images are displayed for context.}
\label{fig:qualitative_comparison}
\end{figure*}
\noindent\textbf{Qualitative Rerults}
Fig.~\ref{fig:qualitative_comparison} illustrates the robustness of our method in highway scenarios under challenging foggy weather and lighting conditions.

\noindent\textbf{Ablation Study} \label{sec:ablation}
To isolate the contribution of each component, we add them incrementally on the baseline MCTrack~\cite{wangMCTrackUnified3D2024}, as shown in Table~\ref{tab:ablation}. Radar Kalman filter refinement achieves the lowest false positives and reduces \gls{ids}. Adding radar association captures more true positives but also increases false positives. This trade-off is stabilized by the cross-check association, which boosts additional true positives, highlighting that robust association is key before refinement. 

\begin{table}[!htbp]
    \caption[Ablation study on \gls{truckscenes}]{Ablation study on \gls{truckscenes} validation set. Each row adds one component on
    top of the MCTrack~\cite{wangMCTrackUnified3D2024} baseline. Bold indicates best results.}
    \label{tab:ablation}
    \renewcommand{\arraystretch}{1.1}
    \centering
    \resizebox{\columnwidth}{!}{%
    \begin{tabular}{ccc ccccc}
    \toprule
    \multicolumn{3}{c}{\textbf{Component}} &
    \multirow{2}{*}{\textbf{AMOTA\%\,$\uparrow$}} &
    \multirow{2}{*}{\textbf{TP\,$\uparrow$}} &
    \multirow{2}{*}{\textbf{FP\,$\downarrow$}} &
    \multirow{2}{*}{\textbf{FN\,$\downarrow$}} &
    \multirow{2}{*}{\textbf{IDS\,$\downarrow$}} \\
    \cmidrule(lr){1-3}
    \makecell{\small Radar\\Kalman filter} &
    \makecell{\small Radar\\Association} &
    \makecell{\small Cross-Check\\Association} & & & & \\
    \midrule
    & &  & 26.6 & 33636 & 13026 & 32378 & 5325 \\
    \checkmark & & & 29.2 & 33948 & 12126 & 32234 & 5157 \\
    \checkmark & \checkmark & & 30.7 & 38446 & 12861 & 26515 & 6378 \\
    \checkmark & \checkmark & \checkmark & \textbf{33.3} & \textbf{42717} & \textbf{12257} & \textbf{24906} & \textbf{3716} \\
    \bottomrule
    \end{tabular}
    }
\end{table}

\section{Conclusion}\label{sec:conclusion}
We present a direct radar fusion method for \gls{3dmot} which does not depend on deep learning. It integrates radar data as additional observations and performs robustly under adverse conditions and at long ranges. As the first \gls{3dmot} study conducted on the \gls{truckscenes} dataset, we release our code to offer a solid baseline for the community. Thanks to the raw data fusion strategy, RadarMOT is practically suitable for robotics and autonomous driving applications.

\bibliographystyle{IEEEtran}
\bibliography{references}
\end{document}